\pdfoutput=1
\documentclass[11pt]{article}

\usepackage[preprint]{acl}

\usepackage{times}
\usepackage{latexsym}
\usepackage{booktabs}
\usepackage{listings}
\usepackage{wrapfig}
\usepackage{subcaption}
\usepackage{graphicx}
\usepackage{booktabs}
\usepackage{pifont}
\usepackage{multirow}
\usepackage{caption}
\newcommand{\cmark}{\textcolor{green!60!black}{\ding{51}}}
\newcommand{\xmark}{\textcolor{red}{\ding{55}}}
\newcommand{\pmark}{\textcolor{orange}{\ding{51}}}  
\usepackage[accsupp]{axessibility}  %
\usepackage[T1]{fontenc}

\usepackage[utf8]{inputenc}

\usepackage{microtype}
\usepackage{url}

\usepackage{inconsolata}

\usepackage{graphicx}
\usepackage{fontawesome5}

\title{UniReason-Med: A Shared Grounded Reasoning Interface for 2D-to-3D Transfer in Medical VQA}

\author{
{\bfseries
Mengzhuo Chen\thanks{Equal contribution} \quad
Yan Shu\footnotemark[1] \quad
Chi Liu\footnotemark[1]
} \\
{\bfseries
Hongming Piao \quad
Xidong Wang \quad
Derek Li \quad
Bryan Dai\thanks{Corresponding author}
} \\
IQuest Research \\
\texttt{\{mzchen, yshu, cliu04, cbdai\}@iquestlab.com}
}

\begin{document}

\maketitle

\vspace*{0.5cm}

\begin{abstract}
We study whether grounded reasoning supervision from abundant 2D medical images can improve 3D medical VQA when both input types are aligned through a common reasoning interface. We introduce UniReason-Med, a single-checkpoint framework that processes either a 2D image or a slice-serialized 3D volume at inference time, generating interleaved textual reasoning and localized visual evidence through shared box syntax, region-token injection, and a common grounded reasoning policy. To train this interface, we construct UniMed-CoT, a 220K instruction-tuning dataset with interleaved textual reasoning and grounded visual evidence, including 170K 2D and 50K 3D samples. Through supervised fine-tuning followed by outcome-level reinforcement learning, UniReason-Med learns to generate grounded reasoning traces without IoU/Dice-based localization rewards during RL. Data-mixture and component ablations show that joint 2D+3D grounded supervision substantially improves 3D reasoning over 3D-only training, while grounding and region-token injection consistently benefit both 2D and 3D tasks. These results suggest that a shared grounded reasoning interface can transfer reasoning structure from 2D images to slice-serialized volumetric medical understanding. The code and data are publicly available at \url{https://github.com/IQuestLab/unireason-med}.
\end{abstract}

\section{Introduction}

Multimodal Large Language Models (MLLMs) have demonstrated remarkable capabilities in visual understanding and text generation, offering a promising foundation for intelligent medical AI systems. By coupling strong language priors with visual perception, MLLMs are increasingly being explored as clinical assistants for medical image interpretation, disease analysis, report generation, and diagnostic support~\cite{li2023llava,chen2024huatuogpt,mullappilly2024bimedix2,lin2025healthgpt}.

However, real-world clinical decision-making poses challenges that extend far beyond general visual understanding. In routine practice, physicians interpret heterogeneous medical data spanning multiple imaging modalities, such as 2D chest X-rays for pneumonia assessment and 3D CT volumes for tumor localization, requiring the synthesis of evidence across fundamentally different spatial representations. This setting calls for a common reasoning interface that can ground and reference evidence consistently across planar images and slice-serialized volumetric scans. Moreover, expert radiologists do not merely inspect images holistically. They ground specific visual findings such as lesions, fractures, or anatomical landmarks and then reason over these localized observations to reach diagnostic conclusions~\cite{rcr2025reporting}. Thus, for universality and interpretability, medical MLLMs should support explicit evidence grounding and cross-dimensional transfer between planar and slice-serialized volumetric reasoning.

Despite recent progress, existing medical MLLMs still fall short of these requirements. Early medical MLLMs, such as LLaVA-Med~\cite{li2023llava} and HuatuoGPT-Vision~\cite{chen2024huatuogpt}, primarily focus on 2D image understanding and instruction following. More recent reasoning-oriented models, including Med-R1~\cite{lai2025med} and MedVLM-R1~\cite{pan2025medvlm}, improve clinical reasoning but operate with only textual chain-of-thought, without explicitly incorporating grounded visual evidence into intermediate reasoning steps. 

\begin{figure*}[t]
\centering
\setlength{\tabcolsep}{6pt} 

\begin{minipage}[t]{0.45\textwidth}
    \centering
    \vspace{0pt}
    \vspace{2.3mm}
    \resizebox{\textwidth}{!}{
    \begin{tabular}{l|c|c|c|c|c|c}
    \toprule
    \multirow{2}{*}{\textbf{Method}} & \multicolumn{2}{c|}{\textbf{Modality}} & \multicolumn{2}{c|}{\textbf{CoT}} & \multicolumn{2}{c}{\textbf{Grounded CoT}} \\
    \cmidrule{2-3} \cmidrule{4-5} \cmidrule{6-7}
    & \textbf{2D} & \textbf{3D} & \textbf{2D} & \textbf{3D} & \textbf{2D} & \textbf{3D} \\
    \midrule
    LLaVA-Med & \cmark & \xmark & \xmark & \xmark & \xmark & \xmark \\
    HuatuoGPT-Vision & \cmark & \xmark & \xmark & \xmark & \xmark & \xmark \\
    RadFM & \cmark & \cmark & \xmark & \xmark & \xmark & \xmark \\
    MedBLIP & \xmark & \cmark & \xmark & \xmark & \xmark & \xmark \\
    M3D-LaMed & \xmark & \cmark & \xmark & \xmark & \xmark & \xmark \\
    Med-R1 & \cmark & \xmark & \cmark & \xmark & \xmark & \xmark \\
    MedVLM-R1 & \cmark & \xmark & \cmark & \xmark & \xmark & \xmark \\
    MAIRA-2 & \cmark & \xmark & \xmark & \xmark & \pmark & \xmark \\
    3DReasonKnee & \xmark & \cmark & \xmark & \cmark & \xmark & \cmark \\
    \midrule
    \textbf{Ours} & \cmark & \cmark & \cmark & \cmark & \cmark & \cmark \\
    \bottomrule
    \end{tabular}
    }
\end{minipage}
\hfill
\begin{minipage}[t]{0.5\textwidth}
    \centering
    \vspace{2mm}
    \includegraphics[width=\textwidth]{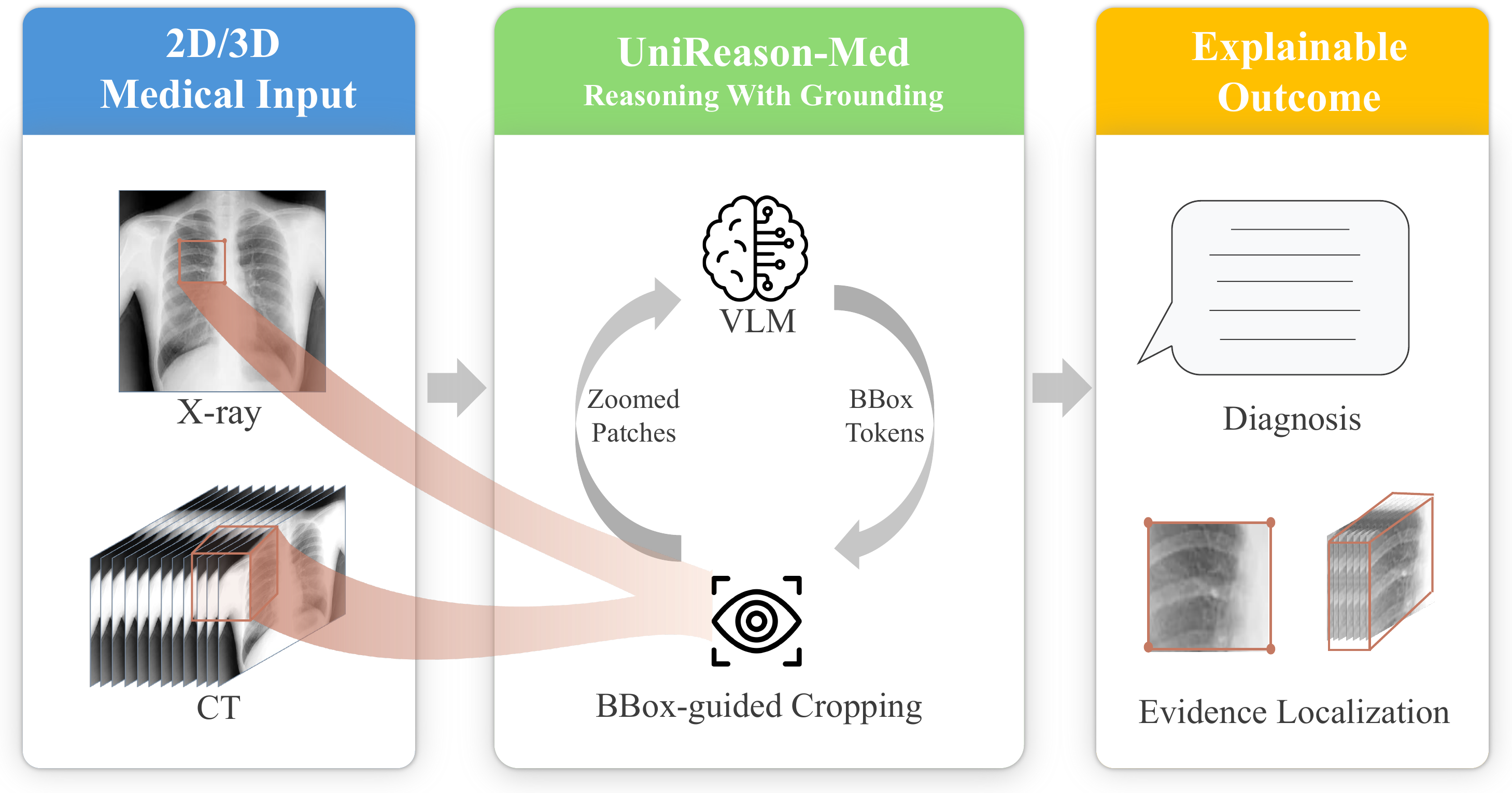}
\end{minipage}

\caption{\textbf{Overview of UniReason-Med.} \textbf{Left:} Comparison of medical MLLMs across key capabilities (\cmark~full support, \pmark~partial, \xmark~none). Checkmarks indicate whether a method reports the corresponding interface capability under public benchmark settings, not clinical readiness or native volumetric representation. UniReason-Med studies a shared grounded reasoning interface for both 2D images and slice-serialized 3D volumes. \textbf{Right:} Each inference instance contains either a 2D image or a 3D volume; the unification lies in the shared language-side box syntax, grounded reasoning policy, and region-token injection mechanism.}
\label{fig:overview}
\vspace{-10pt}
\end{figure*}

In parallel, several works have extended medical MLLMs to 3D imaging or unified 2D-3D modeling, such as RadFM \cite{wu2025towards}, M3D~\cite{bai2024m3d}, OmniV-Med~\cite{jiang2025omnivmed}, and VILA-M3~\cite{nath2025vilam3}. While these methods improve modality coverage and general visual understanding, they mainly emphasize representation unification or task generalization, rather than a reasoning process that explicitly interleaves localized visual evidence with language for either 2D or 3D inputs under a shared interface. On the other hand, grounded medical reasoning has begun to emerge in specialized settings, including grounded report generation and region-aware reasoning on 2D images~\cite{maira2,wang2025v2t,liu2025gemex,leduc2025schain}, as well as recent progress on medical grounding with reinforcement learning and task-specific 3D grounded reasoning datasets~\cite{xu2025medgroundr1,sambara2025threedreasonknee}. Nevertheless, these approaches remain limited either to 2D scenarios, to specific anatomical sites, or to settings where grounding is not seamlessly integrated into the reasoning process. A shared interface for token-interleaved grounded reasoning over both 2D and 3D medical images remains underexplored (Fig.~\ref{fig:overview}).

To address these challenges, we study \textbf{cross-dimensional grounded transfer}: whether abundant 2D grounded reasoning supervision can improve 3D medical VQA when 2D and 3D data share a common reasoning interface. We introduce \textbf{UniReason-Med}, a single-checkpoint framework that processes either a 2D image or a slice-serialized 3D volume while sharing the language model, grounding syntax, and grounded reasoning policy across dimensions. At the core of UniReason-Med is a \textbf{Grounded Chain-of-Thought (GCoT)} interface~\cite{wu2025gcot}, in which reasoning trajectories interleave textual reasoning tokens with visual tokens extracted from regions specified by self-generated 2D boxes or 3D cuboids over ordered slice sequences.

\noindent\textbf{Scope of unification.} We study interface-level unification: 2D images and slice-serialized 3D CT volumes share the same grounding syntax, GCoT policy, region-token injection mechanism, and training objective. Each inference instance contains either a 2D image or a 3D CT volume serialized into 32 ordered slices following M3D~\cite{bai2024m3d}. We evaluate cross-dimensional transfer through joint training under this shared grounded reasoning interface.

To support training, we construct \textbf{UniMed-CoT}, a 220K instruction-tuning dataset with interleaved grounded reasoning annotations generated through an automated pipeline. UniMed-CoT contains 170K 2D and 50K 3D samples spanning diverse imaging modalities and anatomical systems. We first perform supervised fine-tuning on UniMed-CoT to establish the interleaved reasoning format, and then further optimize the model with Group Relative Policy Optimization (GRPO) to improve reasoning quality and grounding consistency~\cite{guo2025deepseek,xu2025medgroundr1}. This design lets us test whether a shared grounded reasoning interface can transfer reasoning structure from abundant 2D supervision to slice-serialized volumetric medical understanding.

In summary, our contributions are fourfold:
\begin{itemize}
\item We formulate cross-dimensional grounded transfer for medical VQA: whether abundant 2D grounded reasoning supervision can improve 3D medical reasoning through a shared language-side interface.
\item We introduce \textbf{UniReason-Med}, which shares box syntax, region-token injection, and a grounded reasoning policy across 2D images and slice-serialized 3D volumes.
\item We construct \textbf{UniMed-CoT}, a 220K-sample dataset with 170K 2D and 50K 3D interleaved grounded reasoning annotations, and validate its quality through automatic filtering and manual inspection.
\item Data-mixture and component ablations show that joint 2D+3D training improves 3D VQA over 3D-only training, grounded visual token injection benefits both dimensions, and outcome-level RL improves grounding without IoU/Dice-based localization rewards during RL.
\end{itemize}

\section{Related Works}

\noindent\textbf{Medical Multimodal Large Language Models.} The rapid evolution of large language models (LLMs) and their multimodal counterparts (MLLMs)~\cite{wang2023huatuo, li2024llava, zhu2025internvl3} has catalyzed progress in medical visual understanding. To adapt general MLLMs to the medical domain, models like LLaVA-Med~\cite{li2023llava}, HuatuoGPT-Vision~\cite{chen2024huatuogpt}, and BiMediX2~\cite{mullappilly2024bimedix2} curate specialized multimodal datasets for pre-training and instruction-tuning. Recently, inspired by DeepSeek-R1~\cite{guo2025deepseek}, efforts including Med-R1~\cite{lai2025med} and Lingshu~\cite{xu2025lingshu} incentivize medical reasoning via verbal chain-of-thought (CoT) and reinforcement learning.

To overcome the limitations of 2D-only models, subsequent works explore 3D and unified 2D--3D architectures. Early generalist models like RadFM~\cite{wu2025towards} and M3D~\cite{bai2024m3d} enable joint processing of 2D and 3D scans. Recent frameworks, such as MedMD~\cite{wu2025towards}, OmniV-Med~\cite{jiang2025omnivmed}, and VILA-M3~\cite{nath2025vilam3}, further unify heterogeneous medical data and enhance 2D--3D information fusion. Despite these advances, a shared medical reasoning interface that consistently grounds and references localized visual evidence for either 2D or 3D inputs remains underexplored.

\noindent\textbf{Visual Chain-of-Thought.} Distinct from textual CoT, visual CoT integrates visual representations into the reasoning process. While early approaches rely on external visual tools (e.g., cropping or zooming)~\cite{zheng2025deepeyes, wang2025pixel}, recent paradigms internalize this mechanism by incorporating localized visual evidence directly into intermediate reasoning steps to reduce hallucinations~\cite{fan2025grit, chen2025mint, wu2025gcot}.

In the medical domain, V2T-CoT~\cite{wang2025v2t} and MAIRA-2~\cite{maira2} explore region-level attention and grounded generation, while S-Chain~\cite{leduc2025schain} introduces structured V-CoT with explicit bounding boxes. Concurrently, MedGround-R1~\cite{xu2025medgroundr1} adapts GRPO to medical grounding without requiring manual CoT annotations. However, these methods are largely limited to 2D images, focus primarily on report generation, or lack interleaved grounded visual tokens during decoding. For 3D imaging, 3DReasonKnee~\cite{sambara2025threedreasonknee} introduces grounded reasoning for knee MRIs, but remains restricted to a single anatomical site. In contrast, our work studies a shared-interface 2D--3D medical reasoning framework that interleaves grounded visual tokens with textual reasoning under a common box representation, enabling controlled analysis of cross-dimensional transfer across diverse imaging modalities and anatomical structures.
\section{Method}
\label{sec:method}

\begin{figure*}[t] 
    \centering
    \includegraphics[width=0.98\textwidth]{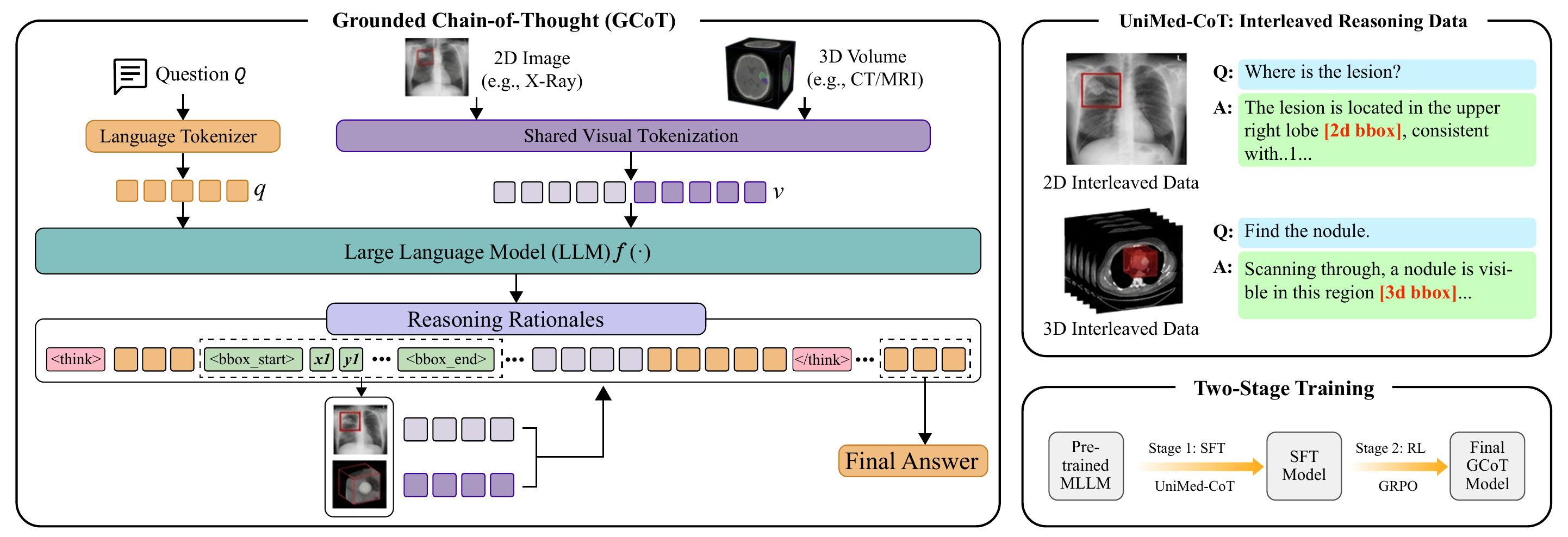}
    \vspace{-0.2cm} 
    \caption{\textbf{Overview of UniReason-Med.}
    (a) Grounded visual evidence extraction for a 2D image or 32-slice CT sequence under the shared GCoT interface.
    (b) Interleaved UniMed-CoT data format.
    (c) Two-stage SFT+GRPO training.}
    \label{fig:framework}
    \vspace{-0.3cm} 
\end{figure*}

In this section, we present the core components of UniReason-Med, a shared-interface framework for grounded visual reasoning over either 2D or 3D medical images. We detail the Grounded Chain-of-Thought (GCoT) interface (Sec.~\ref{sec:gcot}), describe our instruction-tuning dataset UniMed-CoT (Sec.~\ref{sec:dataset}), and introduce our two-stage training paradigm (Sec.~\ref{sec:training}).

\subsection{Grounded Chain-of-Thought (GCoT)}
\label{sec:gcot}

UniReason-Med supports either 2D or 3D medical reasoning at inference time with one shared set of language-side parameters. The model is unified at the token and reasoning-interface level: each instance contains either a 2D image or a 3D CT volume serialized into ordered slices, and both input types share the same grounded output syntax, autoregressive decoder, and GCoT policy.

 Traditional MLLMs $\theta$ output an answer via language-only reasoning:
\begin{equation}
\small
    [\mathbf{r}_1, \mathbf{r}_2, \ldots, \mathbf{r}_k, \mathbf{a}] \sim P_\theta(\cdot \mid I, Q),
\end{equation}
where $I$ denotes image input, $Q$ denotes the text input, $k$ is the number of reasoning steps, $\mathbf{r}_i$ denotes the $i$-th textual reasoning step, and $\mathbf{a}$ is the final answer. However, such purely textual reasoning lacks explicit grounding in visual evidence, limiting both accuracy and interpretability.

To address this, we adapt Grounded Chain-of-Thought (GCoT) to 2D/3D medical reasoning, using a shared interface where the model derives answers from multimodal reasoning chains that interleave textual tokens with visual evidence from grounded regions. Formally, GCoT generates:
\begin{equation}
\small
\begin{aligned}
    &[\mathbf{r}_1, (\mathbf{b}_{1}, \mathbf{v}_{1}), \mathbf{r}_2, (\mathbf{b}_{2}, \mathbf{v}_{2}), \ldots, \\
    &\quad \mathbf{r}_k, (\mathbf{b}_{k}, \mathbf{v}_{k}), \mathbf{a}]
    \sim P_\theta(\cdot \mid I, Q),
\end{aligned}
\end{equation}
where at each reasoning step $i$, the model generates a textual reasoning segment $\mathbf{r}_i$, produces grounding coordinates $\mathbf{b}_i$, and incorporates extracted visual tokens $\mathbf{v}_i$ from the grounded region. The final answer $\mathbf{a}$ is derived based on this multimodal reasoning chain. Crucially, the model learns when and where to ground without relying on an external detector or segmenter at inference time.

At each reasoning step, the model generates coordinates to localize task-relevant regions.
For 2D images, the model outputs bounding box coordinates
$\mathbf{b} = (x_1, y_1, x_2, y_2)$.
For 3D images, we first serialize the volume into an ordered sequence of 2D slices, and the model outputs cuboid coordinates
$\mathbf{b} = (x_1, y_1, z_1, x_2, y_2, z_2)$,
where the $z$ axis can be interpreted as the index of the image slice.
In both cases, the coordinates are represented in absolute image coordinates.

To ensure consistency of the coordinate system across inputs, we apply a smart resize~\cite{bai2025qwen25vltechnicalreport} strategy so that the height and width of each image are divisible by the vision encoder patch size. Following M3D~\cite{bai2024m3d}, each 3D CT volume is uniformly serialized into 32 ordered slices before visual encoding. In implementation, we instantiate the visual encoder with the frozen Qwen2.5-VL vision tower. Thus, our contribution is not a native 3D volumetric encoder, but a $z$-aware grounded reasoning interface over slice-serialized volumetric inputs. Under these preprocessing steps, both 2D and 3D grounding coordinates are consistently defined as absolute positions on the preprocessed input grid.

Given grounding coordinates $\mathbf{b}_i$ at step $i$, we inject focused visual context into the reasoning chain by cropping and encoding the localized region:
\begin{equation}
\small
    X = S_d(I), \quad d \in \{2D, 3D\},
\end{equation}
\begin{equation}
\small
    \mathbf{v}_i = g(f_V(\text{Crop}_d(X, \mathbf{b}_i))),
\end{equation}
where $S_{2D}$ denotes standard 2D image preprocessing and $S_{3D}$ serializes a 3D CT volume into 32 ordered slices following M3D~\cite{bai2024m3d}. The shared frozen vision tower and projector are denoted by $f_V$ and $g$. $\text{Crop}_{2D}$ extracts a 2D rectangular region, whereas $\text{Crop}_{3D}$ extracts the slice range $[z_1,z_2]$ and crops the corresponding $(x,y)$ region on each selected slice. These visual tokens provide evidence that grounds subsequent reasoning in actual image content, enabling the model to reason over what it ``sees'' rather than relying solely on textual descriptions.

\begin{figure*}[t]
    \centering
    \includegraphics[width=0.8\textwidth]{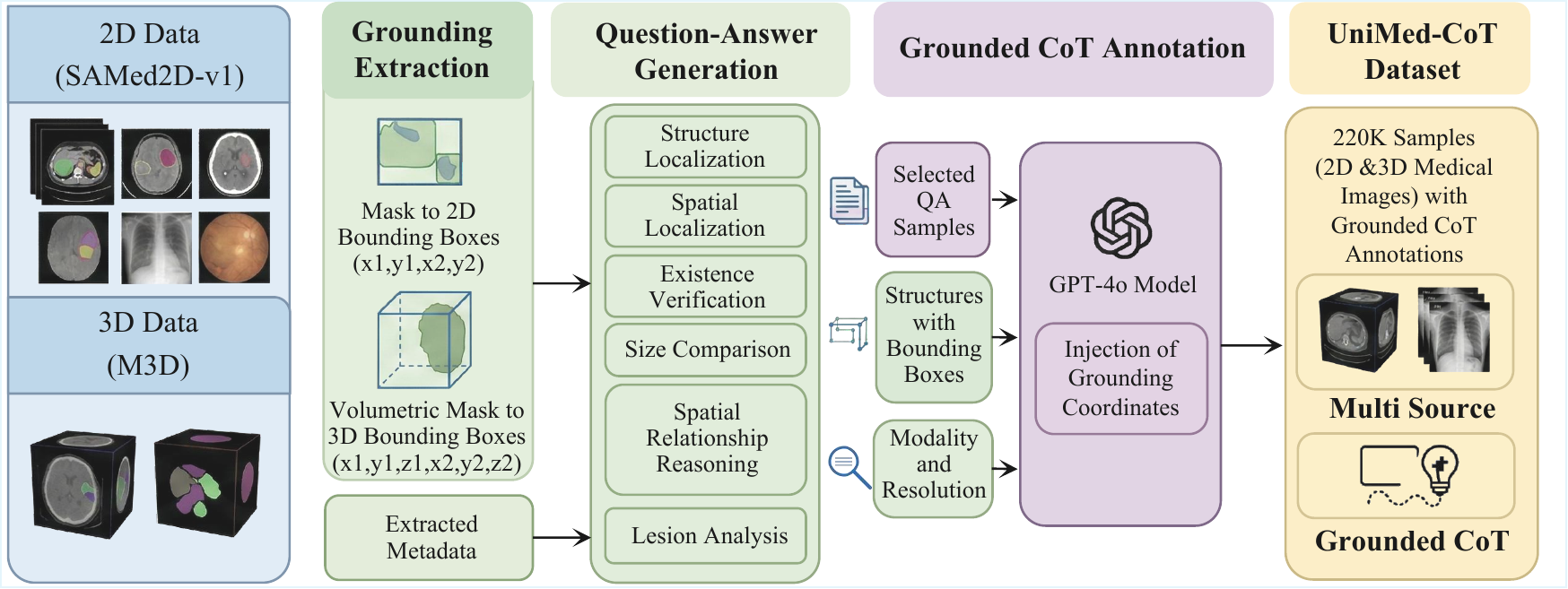}
    \vspace{-0.2cm} 
    \caption{\textbf{UniMed-CoT construction.} From SAMed2D-v1 and M3D segmentation masks, we extract grounding coordinates, generate QA pairs, and use GPT-4o to produce interleaved grounded CoT annotations, yielding 220K 2D/3D samples.}
    \label{fig:dataset}
    \vspace{-0.3cm} 
\end{figure*}

\subsection{UniMed-CoT Dataset}
\label{sec:dataset}

To train UniReason-Med, we construct UniMed-CoT (Fig. \ref{fig:dataset}), a large-scale instruction-tuning dataset comprising 220K grounded chain-of-thought (CoT) samples (170K 2D and 50K 3D) that provide unified supervision across modalities.

\noindent\textbf{Data Sources \& Grounding.} We build upon SAMed2D-v1~\cite{ye2023samed2d20m} for diverse 2D imaging and M3D~\cite{bai2024m3d} training cases for 3D volumetric CTs. Grounding coordinates—$(x_1, y_1, x_2, y_2)$ for 2D and $(x_1, y_1, z_1, x_2, y_2, z_2)$ for 3D—are extracted directly from their existing segmentation masks, ensuring precise spatial grounding without additional annotation overhead. Evaluation images and case identifiers are excluded as summarized in Appendix~\ref{sec:appendix_training}.

\noindent\textbf{QA \& Grounded CoT Generation.} We design question templates covering diverse clinical reasoning dimensions (e.g., spatial localization, relationship reasoning, lesion analysis). We then prompt GPT-4o with the modality, structural boxes, and questions to generate step-by-step reasoning within \texttt{<think>...</think>} blocks. Crucially, we inject grounding coordinates at the first mention of each structure using special tokens: \texttt{<|box\_start|>}\allowbreak\texttt{(x1,y1,x2,y2)}\allowbreak\texttt{<|box\_end|>}\allowbreak\texttt{<region>} for 2D, and \texttt{<|box\_start|>}\allowbreak\texttt{(x1,y1,z1,x2,y2,z2)}\allowbreak\texttt{<|box\_end|>}\allowbreak\texttt{<region>} for 3D. This yields the interleaved reasoning-grounding sequences essential for the GCoT paradigm.

\noindent\textbf{Quality Filtering.} To ensure dataset quality, we discard malformed samples exhibiting missing coordinates, invalid bounding box formats, broken special tokens, or insufficient reasoning depth ($<50$ tokens). After filtering, we retain the final 220K high-quality samples. For the manual audit, we sample 100 2D and 100 3D annotations. Two authors independently check whether (i) the final answer matches the generated question, (ii) the inserted box or cuboid covers the mentioned structure, and (iii) the reasoning text is consistent with the localized evidence; disagreements are resolved by discussion. Overall, 92\% of the inspected annotations exhibit both correct grounding and coherent reasoning, supporting the reliability of the automated pipeline for large-scale grounded-reasoning supervision.

\subsection{Training Paradigm}
\label{sec:training}

We adopt a two-stage training paradigm to establish and subsequently generalize the model's grounded reasoning capabilities.

\paragraph{Stage 1: Grounded CoT SFT.}
In the first stage, we establish foundational grounded reasoning via supervised fine-tuning (SFT) on UniMed-CoT. As introduced in Sec.~\ref{sec:gcot}, each training sequence interleaves textual reasoning segments $\mathbf{r}_i$, grounding coordinates $\mathbf{b}_i$, inserted visual region embeddings $\mathbf{v}_i$, and the final answer $\mathbf{a}$. Let $Y = [y_1, y_2, \ldots, y_T]$ denote the discrete output-token sequence after excluding inserted continuous region embeddings. We apply standard auto-regressive cross-entropy loss over all discrete target tokens $\mathcal{D} \subset \{1, 2, \ldots, T\}$, including reasoning text, special markers, grounding coordinates, and the final answer:
\begin{equation}
\small
    \mathcal{L}_{\text{SFT}} = -\sum_{t \in \mathcal{D}} \log P_\theta(y_t \mid y_{<t}, I, Q),
\end{equation}
where $I$ and $Q$ denote the input image and question. The inserted visual region embeddings $\mathbf{v}_i$ are continuous features from cropped regions and are therefore excluded from token-level supervision. This stage teaches the model to localize relevant regions and incorporate visual evidence during reasoning.

\paragraph{Stage 2: Grounded CoT RL with GRPO.}
To generalize beyond SFT annotations, we apply Group Relative Policy Optimization (GRPO)~\cite{shao2024deepseekmath}. This stage is useful for mitigating overfitting to imperfect automated artifacts in UniMed-CoT and encouraging flexible exploration of grounding strategies.

By shifting from process-level imitation to outcome-level supervision, GRPO refines established reasoning patterns using outcome-driven rewards for valid final solutions. Notably, we deliberately exclude ground-truth box-overlap localization rewards such as IoU or Dice, allowing us to evaluate whether outcome-level rewards can implicitly enhance spatial grounding.

\noindent\textbf{Group sampling and rewards.} For each input $x=(I, Q)$, we sample a group of $G$ reasoning chains $\{y_j\}_{j=1}^{G}$ from the current policy. Each candidate receives a composite reward:
\begin{equation}
\small
    r(y_j) = r^{\text{ans}}(y_j) + \lambda \cdot r^{\text{format}}(y_j),
\end{equation}
where $r^{\text{ans}}, r^{\text{format}} \in \{0, 1\}$ strictly indicate answer correctness and adherence to the valid coordinate format, respectively. Rewards are normalized into relative advantages within the group to stabilize training:
\begin{equation}
\small
    A_j = \frac{r(y_j) - \mu_r}{\sigma_r + \epsilon},
\end{equation}
with $\mu_r$ and $\sigma_r$ representing the group mean and standard deviation, and $\epsilon$ for numerical stability.

\noindent\textbf{GRPO objective.} Let $\pi_\theta$ be the trainable policy and $\pi_{\text{old}}$ the reference policy (initialized from SFT). GRPO optimizes the clipped policy gradient:
\begin{equation}
\small
\begin{split}
    \mathcal{L}_{\text{GRPO}} = &-\mathbb{E}_{x,\{y_j\}} \Bigg[ \frac{1}{G} \sum_{j=1}^{G} \min \Big( \rho_j A_j, \\
    & \text{clip}(\rho_j, 1-\epsilon_c, 1+\epsilon_c) A_j \Big) \Bigg],
\end{split}
\end{equation}
where $\rho_j = \frac{\pi_\theta(y_j \mid x)}{\pi_{\text{old}}(y_j \mid x)}$ and $\epsilon_c$ controls the clipping margin. By leveraging group-wise relative advantages, GRPO stably assigns higher probabilities to high-reward trajectories, unleashing interleaved reasoning without requiring ground-truth box-overlap supervision.
\section{Experiments}
\label{sec:experiment}

Our experiments aim to address three research questions: 
\textbf{RQ1:} Does joint 2D+3D grounded supervision improve 3D medical VQA over 3D-only training? 
\textbf{RQ2:} Which components of the shared grounded reasoning interface contribute to performance? 
\textbf{RQ3:} Does outcome-level RL improve grounding without IoU/Dice-based localization rewards during RL? 

\subsection{Evaluation Benchmarks and Baselines}

\textbf{Benchmarks.} We evaluate cross-modal medical visual understanding on OmniMedVQA~\cite{hu2024omnimedvqa} for 2D scenarios, encompassing multiple modalities (e.g., CT, MRI, X-ray, fundus). For 3D evaluation, we utilize M3D-VQA~\cite{bai2024m3d}, which challenges models on clinically relevant tasks including spatial reasoning, abnormality detection, and organ recognition within slice-serialized volumetric scans.

\textbf{Baselines.} We compare our method against representative general-domain and medical-specific multimodal large language models (MLLMs). For 2D VQA, baselines include generalist models (e.g., MiniGPT-4, InstructBLIP, LLaVA, Qwen2.5VL) and medical specialists (e.g., LLaVA-Med, MedPLIB). We additionally report recent frontier/reference models where reproducible evaluation is available to contextualize the gap between compact open-source medical models and much larger proprietary systems. For 3D evaluation, we compare against recent volumetric MLLMs, including M3D-LaMed~\cite{bai2024m3d}, Lingshu~\cite{xu2025lingshu}, and MedGemma~\cite{medgemma}. For all Qwen2.5-VL-based models on M3D-VQA, including the backbone baselines and UniReason-Med, 3D CT volumes are serialized into the same 32-slice input sequence following the M3D protocol.

\begin{table*}[t]
\small
\centering

\caption{Data-mixture ablation during SFT under the same optimization schedule. Joint 2D+3D grounded-reasoning supervision improves 3D VQA over 3D-only training with matched training steps.}
\vspace{-10pt}
\label{tab:boost_ablation}

\begin{tabular}{l|cccccc}
\toprule
\textbf{Model} & \textbf{Plane} & \textbf{Phase} & \textbf{Organ} & \textbf{Abnormality} & \textbf{Location} & \textbf{Mean} \\
\midrule
Qwen2.5VL-7B & 91.3 & 45.5 & 47.7 & 48.9 & 53.1 & 57.3 \\
UniReason-Med (2D-only) & 90.7 & 38.6 & 52.4 & 46.1 & 45.2 & 56.2 \\
UniReason-Med (3D-only) & 90.1 & 31.6 & 65.3 & 63.5 & 55.8 & 61.2 \\
UniReason-Med (2D+3D)& \textbf{97.7} & \textbf{60.2} & \textbf{69.7} & \textbf{63.6} & \textbf{60.0} & \textbf{70.2} \\
\bottomrule
\end{tabular}

\end{table*}
\begin{table*}[t]
\small
\centering
\caption{Contextual performance on the 2D benchmark OmniMedVQA. We additionally report representative recent frontier/reference models where reproducible evaluation is available.}
\label{tab:2dvqa}
\vspace{-10pt}
\begin{tabular}{l|c|cccccccc|c}
\toprule
\textbf{Model} & \textbf{Param.} & \textbf{CT} & \textbf{MR} & \textbf{OCT} & \textbf{Der} & \textbf{MIC} & \textbf{X-Ray} & \textbf{FP} & \textbf{US} & \textbf{Mean} \\
\midrule
MiniGPT-4  & 7B & 23.7 & 28.7 & 33.6 & 41.3 & 29.3 & 37.2 & 42.5 & 26.4 & 32.8 \\
BLIP-2  & 4B & 59.9 & 43.5 & 69.6 & 40.9 & 51.5 & 65.0 & 67.6 & 39.1 & 54.6 \\
InstructBLIP  & 7B & 29.5 & 36.1 & 45.5 & 63.0 & 48.7 & 58.1 & 44.3 & \textbf{43.4} & 46.1 \\
LLaVA  & 7B & 18.3 & 28.9 & 37.2 & 49.7 & 37.8 & 28.4 & 34.1 & 23.2 & 31.0 \\
VPGTrans & 7B & 22.9 & 26.5 & 27.2 & 45.4 & 25.5 & 44.2 & 36.8 & 27.5 & 32.0 \\
RadFM  & 14B & 27.9 & 24.7 & 34.0 & 38.3 & 26.3 & 26.6 & 31.4 & 16.5 & 28.2 \\
LLaVA-Med  & 7B & 19.6 & 30.5 & 39.0 & 46.4 & 29.3 & 32.4 & 43.1 & 30.4 & 33.8 \\
LISA  & 7B & 63.0 & 49.0 & 66.2 & 41.6 & 54.7 & 62.3 & 46.7 & 32.1 & 52.0 \\
MedPLIB  & 12B & 62.7 & 67.0 & 75.1 & 51.5 & 64.4 & 60.3 & 65.0 & 38.8 & 60.6 \\
Qwen2.5VL-7B & 7B & 69.7 & 61.3 & 59.6 & 60.5 & 64.4 & 76.2 & 74.5 & 32.4 & 61.4 \\
Qwen2.5VL-32B & 32B & \textbf{70.5} & 69.7 & 69.1 & 67.2 & 65.9 & 76.0 & 72.4 & 39.1 & 66.0 \\
Qwen3.5-9B & 9B & 72.2 & 68.5 & 69.8 & 68.0 & 68.5 & 77.5 & 75.0 & 42.9 & 67.8 \\ 
GPT-5 & -- & 85.2 & 81.5 & 86.8 & 84.0 & 80.5 & 88.5 & 89.0 & 64.5 & \textbf{82.5} \\ 
\midrule
\textbf{UniReason-Med} & 7B & 69.7 & \textbf{78.9} & \textbf{82.2} & \textbf{70.9} & \textbf{66.9} & \textbf{81.3} & \textbf{85.3} & 34.2 & \textbf{71.1} \\
\bottomrule
\end{tabular}
\end{table*}
\subsection{Training Setup}

We adopt Qwen2.5-VL-7B-Instruct as our base model and employ a two-stage training pipeline. Detailed hyperparameters, system configurations, and the data-isolation summary for both stages are provided in Appendix~\ref{sec:appendix_training}. We enforce image-identifier, case-identifier, metadata, hash, and image-similarity checks where available to reduce leakage and near-duplicate contamination.

\paragraph{Stage 1: Supervised Fine-Tuning (SFT).} 
We optimize only the language model on our UniMed-CoT corpus (170K 2D samples from SAMed2D-v1; 50K 3D samples from M3D), leaving the vision tower and projector frozen. For the SFT data-mixture ablations, each mixture is trained with the same update-step budget via resampling when needed. Strict data decontamination ensures no overlap with OmniMedVQA or M3D-VQA test splits.

\paragraph{Stage 2: Reinforcement Learning (RL).} 
Following SFT, we further optimize the reasoning policy via GRPO. To construct the RL environment, we sample 40K instances equally distributed between the PMC-VQA~\cite{zhang2023pmc} training split and the M3D-VQA~\cite{bai2024m3d} training split. These RL instances are disjoint from the SFT corpus and exclude OmniMedVQA/M3D-VQA test images or cases. During RL, the model explores diverse reasoning trajectories using group sampling, driven by the composite outcome-based reward described in Sec.~\ref{sec:training}.

\begin{table*}[t]
\small
\centering
\caption{Performance on the 3D benchmark M3D-VQA.}
\label{tab:3d_vqa}
\vspace{-4pt}
\begin{tabular}{l|cccccc}
\toprule
\textbf{Model} & \textbf{Plane} & \textbf{Phase} & \textbf{Organ} & \textbf{Abnormality} & \textbf{Location} & \textbf{Mean} \\
\midrule
RadFM & 19.7 & 28.7 & 16.8 & 18.9 & 14.9 & 19.8 \\
Qwen2.5VL-7B & 91.3 & 45.5 & 47.7 & 48.9 & 53.1 & 57.3 \\
Qwen2.5VL-32B & 90.8 & 49.4 & 53.7 & 57.4 & 54.8 & 61.2 \\
MedGemma-4B-IT & 75.5 & 32.3 & 51.0 & 58.8 & 51.3 & 53.8 \\
LLaVA-Med-7B & 73.4 & 18.4 & 39.0 & 41.8 & 30.6 & 40.6 \\
M3D-LaMed-Llama-2-7B & 98.8 & 79.8 & 74.8 & 66.7 & 58.9 & 75.8 \\
M3D-LaMed-Phi-3-4B & 98.6 & 87.4 & 76.8 & 74.4 & 62.4 & 79.9 \\
Lingshu-7B & 97.3 & 61.1 & 57.6 & 63.6 & 55.2 & 66.9 \\
\midrule
UniReason-Med & \textbf{99.0} & \textbf{92.1} & \textbf{81.6} & \textbf{80.3} & \textbf{66.0} & \textbf{83.8} \\
\bottomrule
\end{tabular}
\vspace{-8pt}
\end{table*}

\begin{table}[t]
\centering
\small

\caption{Ablation of the shared GCoT interface. Adding coordinates (Box-only) and cropped visual tokens (Full) improves both 2D and 3D VQA.}
\vspace{-4pt}
\begin{tabular}{lcc}
\toprule
\textbf{Model Setup} & \textbf{2D Mean} & \textbf{3D Mean} \\
\midrule
Text-CoT SFT & 62.0 & 67.7 \\
Box-only GCoT SFT & 63.7 & 69.7 \\
\textbf{Full GCoT SFT} & \textbf{65.3} & \textbf{70.2} \\
\bottomrule
\end{tabular}
\label{tab:gcot_ablation}
\end{table}

\begin{table}[t]
\caption{Ablation study on training stages.}
\vspace{-4pt}
\centering
\small
\begin{tabular}{lcc}
\toprule
\textbf{Model} & \textbf{2D Mean} & \textbf{3D Mean} \\
\midrule
Qwen2.5VL-7B & 61.4 & 57.3 \\
UniReason-Med (SFT-only) & 65.3 & 70.2 \\
UniReason-Med (RL-only) & 66.8 & 69.3 \\
\textbf{UniReason-Med (SFT+RL)} & \textbf{71.1} & \textbf{83.8} \\
\bottomrule
\end{tabular}

\label{tab:stage_ablation}
\vspace{-8pt}
\end{table}

\begin{table}[t]
\centering
\small
\caption{Grounding evaluation (Dice). RL improves grounding on 2D and 3D datasets without IoU/Dice localization rewards during RL.}
\vspace{-4pt}
\begin{tabular}{llcc}
\toprule
\textbf{Dim.} & \textbf{Dataset} & \textbf{SFT} & \textbf{SFT+RL} \\
\midrule
2D & Kvasir-SEG & 0.54 & \textbf{0.65} \\
3D & AMOS22 & 0.36 & \textbf{0.42} \\
3D & MSD & 0.38 & \textbf{0.45} \\
\bottomrule
\end{tabular}
\label{tab:grounding_eval}
\vspace{-8pt}
\end{table}

\subsection{Experiments}

\paragraph{\textbf{RQ1:} Does joint 2D+3D grounded supervision improve 3D medical VQA over 3D-only training?}

\textbf{Cross-Dimensional Transfer.} We investigate whether abundant 2D grounded reasoning data can enhance 3D capabilities via our shared GCoT interface. As shown in Table~\ref{tab:boost_ablation}, 2D-only training degrades 3D performance, confirming that cross-modal transfer strictly requires 3D anchors. However, joint 2D+3D training boosts the 3D mean accuracy from 61.2 (3D-only) to 70.2. Since all SFT ablations share the same update-step budget and compute through mixture resampling, the gain is not explained by additional optimization steps alone; it is associated with adding 2D grounded supervision to 3D anchors under the shared GCoT interface. These results support our core hypothesis: 2D grounded supervision substantially benefits slice-serialized volumetric reasoning when aligned with 3D data.

\textbf{Contextual 2D and 3D VQA Performance.} We further benchmark a single unified checkpoint across planar and volumetric datasets to contextualize this learned interface. On the 2D OmniMedVQA (Table~\ref{tab:2dvqa}), UniReason-Med achieves a 71.1 mean accuracy, outperforming its Qwen2.5VL-7B backbone by 9.7 points and the larger 32B variant by 5.1 points, establishing strong open-source performance (though a gap to proprietary frontier models like GPT-5 remains). On the 3D M3D-VQA (Table~\ref{tab:3d_vqa}), UniReason-Med reaches 83.8 mean accuracy, surpassing the strongest open-source baseline, M3D-LaMed-Phi-3-4B, by 3.9 points.

\paragraph{\textbf{RQ2:} Which components of the shared grounded reasoning interface contribute to performance?}

\textbf{(1) GCoT Interface (Table~\ref{tab:gcot_ablation}):} Upgrading from Text-CoT to Box-only GCoT (adding coordinates) and Full GCoT (injecting cropped visual tokens) consistently improves mean accuracy in both 2D (62.0$\rightarrow$63.7$\rightarrow$65.3) and 3D (67.7$\rightarrow$69.7$\rightarrow$70.2). This confirms that explicit spatial and visual grounding benefits both modalities.
\textbf{(2) Training Stages (Table~\ref{tab:stage_ablation}):} While applying SFT or RL individually to the base Qwen2.5VL-7B yields strong gains, combining both stages (SFT+RL) achieves peak performance (71.1 in 2D, 83.8 in 3D). SFT effectively establishes foundational reasoning priors, which RL subsequently refines and generalizes.

\paragraph{\textbf{RQ3:} Does outcome-level RL improve grounding without IoU/Dice-based localization rewards during RL?}

\textbf{Implicit Grounding Improvement.} We conduct a quantitative evaluation of grounding quality. For grounding evaluation, predicted 2D boxes and 3D cuboids are clipped to valid image boundaries, rescaled to the evaluation resolution, converted into binary rectangular or cuboid masks, and compared with ground-truth segmentation masks using per-case Dice. Segmentation masks are used only for evaluation and are not used as RL rewards. As shown in Table~\ref{tab:grounding_eval}, the RL stage consistently improves localization for both planar and slice-serialized volumetric inputs, despite lacking ground-truth box-overlap supervision during RL. On the 2D Kvasir-SEG benchmark~\cite{jha2019kvasir}, Dice improves from 0.54 to 0.65. Similarly, 3D Dice scores increase from 0.36 to 0.42 (AMOS22) and 0.38 to 0.45 (MSD).

\textbf{Correlation between Grounding and Reasoning.} To understand this implicit improvement, we analyze the Pearson correlation between grounding IoU and final answer correctness across all test samples. As detailed in Appendix~\ref{sec:appendix_pearson}, the two metrics exhibit a significant positive correlation ($r = 0.832, p < 1.7\times10^{-52}$). This positive correlation is consistent with the GCoT hypothesis that more accurate localization tends to provide more reliable visual tokens for reasoning. It also suggests a plausible mechanism for why answer-level RL can improve grounding indirectly: optimizing for answer correctness may favor trajectories with more accurate intermediate grounding.

\textbf{Slice-volume Grounding Consistency.} To further examine this shared syntax, we analyze consistency between slice-wise and volume-level grounding. For each generated 3D cuboid, we compare its $xy$ projection within the predicted $z$ range against 2D GCoT predictions on the corresponding slices. The average slice-volume Intersection-over-Union (IoU) improves from 0.42 (SFT-only) to 0.57 (SFT+RL). As visualized in Appendix~\ref{sec:appendix_qualitative}, our shared interface encourages the model to focus on consistent anatomical regions across planar slices and slice-serialized volumetric inputs.

\section{Conclusion}

We introduced UniReason-Med, a shared GCoT interface for studying 2D-to-3D transfer in medical VQA. With UniMed-CoT, a 220K grounded-reasoning dataset, the model interleaves textual reasoning with localized evidence for 2D images and slice-serialized 3D volumes. Experiments show that, under matched SFT steps, joint 2D+3D supervision improves 3D VQA over 3D-only training, while outcome-level RL improves grounding without IoU/Dice rewards.
\section*{Limitations}

Our experiments mainly follow established public benchmarks, which provide reproducible comparison against prior medical MLLMs and grounded reasoning systems. Future work can extend this evaluation with prospective radiologist studies and downstream clinical workflows. UniReason-Med follows the M3D-style 32-slice serialization protocol for 3D inputs; extending the same grounded reasoning interface to denser volumetric encoders is a natural next step. UniMed-CoT is built with an automated GPT-4o-based annotation pipeline, and our filtering, manual spot checks, and outcome-based RL are designed to improve the reliability of the resulting grounded reasoning traces.

\section*{Ethical Considerations and Responsible Use}

UniReason-Med is intended for research use and benchmark analysis, not for autonomous clinical decision-making. Incorrect answers or inaccurate grounding could mislead downstream users in high-stakes diagnostic settings; deployment would require prospective clinical validation, expert oversight, and institution-specific safety review. We use publicly released, de-identified medical imaging datasets and do not collect new patient-identifying information, patient names, or free-form clinical notes. We plan to release our code under the MIT License and the UniMed-CoT annotations, prompts, and metadata that we create under CC BY 4.0, while the underlying images and third-party artifacts remain governed by their original licenses or terms of use. UniMed-CoT and UniReason-Med are intended for research use consistent with those source-data access conditions. ChatGPT was used for limited language polishing during writing; all technical claims, experimental results, and final text were reviewed and edited by the authors.

\bibliography{custom}

\clearpage
\newpage
\appendix
\section{Organization of the Appendix}
This appendix provides supplementary details to ensure the reproducibility and comprehensiveness of our study. \textbf{Appendix~\ref{sec:appendix_training}} details the hyperparameter configurations and our strict data isolation strategy. \textbf{Appendix~\ref{sec:appendix_prompt}} provides the specific prompt templates used for querying GPT-4o to construct the UniMed-CoT dataset. \textbf{Appendix~\ref{sec:appendix_reward}} elaborates on the programmatic implementation of the reward functions used during the GRPO stage. Finally, \textbf{Appendix~\ref{sec:appendix_qualitative}} presents qualitative visual examples of our model's grounded reasoning process.

\section{Detailed Training Configurations}
\label{sec:appendix_training}

To support reproducibility and maintain a concise main text, this section provides the comprehensive training details and hyperparameters for the two-stage training paradigm of UniReason-Med. 

\subsection{Hyperparameter Settings}

Table~\ref{tab:hyperparameters} summarizes the key hyperparameters used during both the Supervised Fine-Tuning (SFT) and Reinforcement Learning (RL) stages. For the SFT data-mixture ablations, we use a fixed update-step budget and resample each mixture when needed so that 3D-only, 2D-only, and joint 2D+3D settings are compared under the same optimization schedule. All training is conducted using bfloat16 precision to optimize memory efficiency while maintaining numerical stability, and is parallelized using the DeepSpeed ZeRO-3 optimization framework.

\begin{table}[h]
\centering
\resizebox{\linewidth}{!}{
\begin{tabular}{lc}
\toprule
\textbf{Hyperparameter} & \textbf{Value} \\
\midrule
\multicolumn{2}{c}{\textit{Stage 1: Supervised Fine-Tuning (SFT)}} \\
\midrule
Base Model & Qwen2.5-VL-7B-Instruct \\
Trainable Parameters & LLM only (Vision \& Projector frozen) \\
Per-device Batch Size & 2 \\
Gradient Accumulation Steps & 4 \\
Effective Batch Size & 8 \\
Learning Rate & $1 \times 10^{-5}$ \\
LR Scheduler & Cosine \\
Warmup Ratio & 0.03 \\
SFT Schedule & Fixed update steps with mixture resampling \\
Precision & bfloat16 \\
DeepSpeed Stage & ZeRO-3 \\
\midrule
\multicolumn{2}{c}{\textit{Stage 2: Reinforcement Learning (GRPO)}} \\
\midrule
Global Training Batch Size & 64 \\
PPO Mini-batch Size & 32 \\
Actor Learning Rate & $1 \times 10^{-6}$ \\
Rollout Samples per Prompt ($G$) & 8 \\
\bottomrule
\end{tabular}
}
\caption{Comprehensive hyperparameter configurations for the two-stage training pipeline.}
\label{tab:hyperparameters}
\end{table}

\subsection{Strict Data Isolation Strategy}

To prevent data leakage and make the training mixture reproducible, Table~\ref{tab:data_isolation} summarizes the data used in each stage. In the reported experiments, Stage-1 SFT uses UniMed-CoT as the grounded-reasoning corpus, consisting of 170K 2D samples derived from SAMed2D-v1 and 50K 3D samples derived from M3D training cases; all Stage-1 SFT sources are listed in the table. Stage-2 RL uses a disjoint 40K-instance mixture sampled from the PMC-VQA and M3D-VQA training splits.

We enforce image-identifier and case-identifier isolation across SFT, RL, and evaluation. All evaluation images/cases from OmniMedVQA and M3D-VQA test splits are excluded from training. For M3D-derived data, we follow the official train/test split and remove any case identifiers overlapping with the M3D-VQA test set. To reduce near-duplicate contamination across repackaged public datasets, we additionally apply available metadata checks, exact-file hashing, perceptual hashing for 2D images and rendered slices, and image-similarity screening for candidate overlaps between training sources and evaluation splits. Kvasir-SEG, AMOS22, and MSD are used only for grounding evaluation; images/cases flagged by identifier, metadata, hash, or similarity checks against these grounding-evaluation splits are removed from UniMed-CoT construction and RL training.

\begin{table*}[t]
\centering
\small
\resizebox{\textwidth}{!}{%
\begin{tabular}{lllll}
\toprule
\textbf{Source} & \textbf{\#Samples} & \textbf{Modality} & \textbf{Stage} & \textbf{Overlap policy / relation to evaluation} \\
\midrule
UniMed-CoT / SAMed2D-v1-derived & 170K & 2D & SFT & Used for grounded-reasoning SFT; OmniMedVQA test images are excluded. \\
UniMed-CoT / M3D-derived training cases & 50K & 3D CT & SFT & Follows the official M3D split; M3D-VQA test case identifiers are excluded. \\
PMC-VQA training split & 20K & 2D & RL & Sampled only for RL; disjoint from SFT and OmniMedVQA test images. \\
M3D-VQA training split & 20K & 3D CT & RL & Sampled only for RL; official test case identifiers are excluded. \\
OmniMedVQA test split & Eval only & 2D & Evaluation & Never used for SFT or RL. \\
M3D-VQA test split & Eval only & 3D CT & Evaluation & Never used for SFT or RL. \\
Kvasir-SEG grounding split & Eval only & 2D & Evaluation & Used only for grounding evaluation; excluded from SFT and RL via identifier/hash/similarity checks. \\
AMOS22 grounding split & Eval only & 3D CT & Evaluation & Used only for grounding evaluation; excluded from SFT and RL via identifier/metadata/hash/similarity checks. \\
MSD grounding split & Eval only & 3D CT & Evaluation & Used only for grounding evaluation; excluded from SFT and RL via identifier/metadata/hash/similarity checks. \\
\bottomrule
\end{tabular}
}
\caption{Training and evaluation data isolation summary.}
\label{tab:data_isolation}
\end{table*}

\section{Prompt Design for Grounded CoT Generation}
\label{sec:appendix_prompt}

To construct the UniMed-CoT dataset with high-quality step-by-step reasoning, we rely on GPT-4o. The prompt is meticulously designed to inject spatial metadata (bounding boxes) into the context, forcing the model to associate medical structures with their corresponding coordinates before generating the final answer. 

Table~\ref{tab:prompt_template} illustrates the unified prompt template used for both 2D and 3D modalities. By providing the exact spatial coordinates of key structures within the prompt, we ensure that GPT-4o's generated reasoning within the \texttt{<think>} blocks is implicitly grounded. During post-processing, we string-match the structure names and append our special coordinate tokens (e.g., \texttt{<|box\_start|>}\allowbreak\texttt{(x1,y1,x2,y2)}\allowbreak\texttt{<|box\_end|>}) at their first mention to construct the final interleaved SFT sequence.

\begin{table}[h]
\centering
\small
\begin{tabular}{p{0.95\linewidth}}
\toprule
\textbf{System Prompt} \\
\midrule
\texttt{You are an expert radiologist and medical AI assistant. You will be provided with the metadata of a medical image, including its modality, resolution, and a list of localized structures with their bounding box coordinates. Given a medical question, you must first think step-by-step in a rigorous clinical manner to deduce the answer. Wrap your reasoning process inside <think> and </think> tags. Then, provide the final answer.} \\
\midrule
\textbf{User Input Template} \\
\midrule
\texttt{\textbf{[Modality]:} \{modality\_type\} (e.g., 3D CT, 2D MRI)} \\
\texttt{\textbf{[Resolution]:} \{width\} x \{height\} (x \{depth\} for 3D)} \\
\texttt{\textbf{[Annotated Structures]:}} \\
\texttt{- \{structure\_1\}: \{box\_coordinates\_1\}} \\
\texttt{- \{structure\_2\}: \{box\_coordinates\_2\}} \\
\texttt{...} \\
\texttt{\textbf{[Question]:} \{generated\_question\}} \\
\bottomrule
\end{tabular}
\caption{The unified prompt template used to query GPT-4o for generating grounded chain-of-thought annotations in the UniMed-CoT dataset.}
\label{tab:prompt_template}
\end{table}

\begin{figure*}[t]
    \centering
    \includegraphics[width=0.95\textwidth]{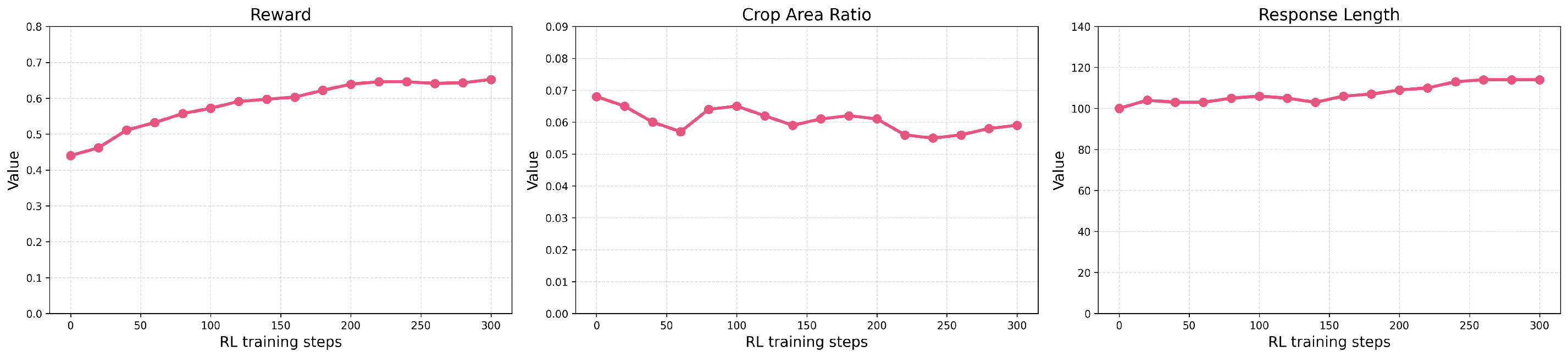} 
    \vspace{-0.2cm}
    \caption{\textbf{Evolution of key metrics during reinforcement learning training.} The reward steadily increases as training progresses, indicating improved policy performance. 
Meanwhile, the crop area ratio shows a slightly decreasing trend with minor fluctuations, and the response length shows a mild upward trend.}
    \label{fig:training_curve}
    \vspace{-0.3cm}
\end{figure*}

\section{Reward Function Implementation in GRPO}
\label{sec:appendix_reward}

During the Stage 2 Reinforcement Learning, our GRPO framework optimizes the policy using a composite reward $r(y_j) = r^{\text{ans}}(y_j) + \lambda \cdot r^{\text{format}}(y_j)$ (where $\lambda = 0.5$). We avoid using ground-truth box-overlap metrics (e.g., IoU or Dice) as localization rewards, aiming to test whether outcome-level supervision naturally induces better intermediate grounding. The rewards are implemented programmatically as follows:

\textbf{Answer Correctness Reward ($r^{\text{ans}}$):} 
This reward assigns a value of $1.0$ if the final answer matches the ground truth, and $0.0$ otherwise. For closed-source multiple-choice or yes/no questions, we extract the generated answer using regular expressions and perform exact string matching. For open-ended questions, we employ a rule-based matching mechanism that checks for the presence of key clinical entities or utilize a lightweight LLM (e.g., Llama-3-8B) as an automated judge to determine semantic equivalence between the prediction and the ground truth.

\textbf{Format Reward ($r^{\text{format}}$):} 
This reward acts as a strict structural constraint to ensure the model produces valid interleaved GCoT sequences. We use Python's \texttt{re} module to verify the following criteria, granting a reward of $1.0$ only if all are satisfied:
\begin{enumerate}
    \setlength{\itemsep}{0pt}
    \item The output must contain exactly one pair of \texttt{<think>} and \texttt{</think>} tags.
    \item The text outside the \texttt{<think>} tags must directly address the final answer.
    \item Any generated coordinates must strictly follow the defined grammatical structure. We use the regex pattern 
\texttt{<|box\_start|>}\allowbreak\texttt{\textbackslash((}\allowbreak\texttt{\textbackslash d+,\textbackslash d+,}\allowbreak\texttt{\textbackslash d+,\textbackslash d+}\allowbreak\texttt{(?:,\textbackslash d+,\textbackslash d+)?}\allowbreak\texttt{)\textbackslash)<|box\_end|>} 
to ensure no malformed coordinates are generated.
\end{enumerate}
If the model hallucinates irregular coordinate formats or fails to close the reasoning tags, it receives a $0.0$ format reward, rapidly penalizing degenerate reasoning trajectories.

\begin{figure}[h]
    \centering
    \includegraphics[width=0.8\linewidth]{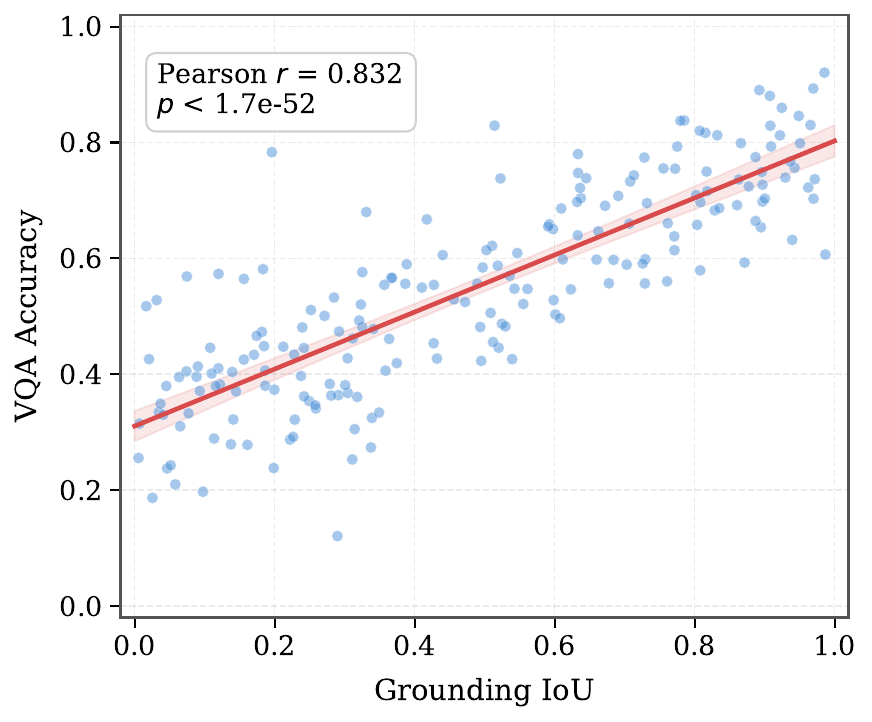}
    \caption{Pearson correlation analysis between grounding IoU and answer correctness ($r = 0.832$).}
    \label{fig:pearson}
\end{figure}
\section{RL Training Dynamics}
\label{sec:appendix_dynamics}

\begin{figure*}[t]
    \centering
    \includegraphics[width=\textwidth]{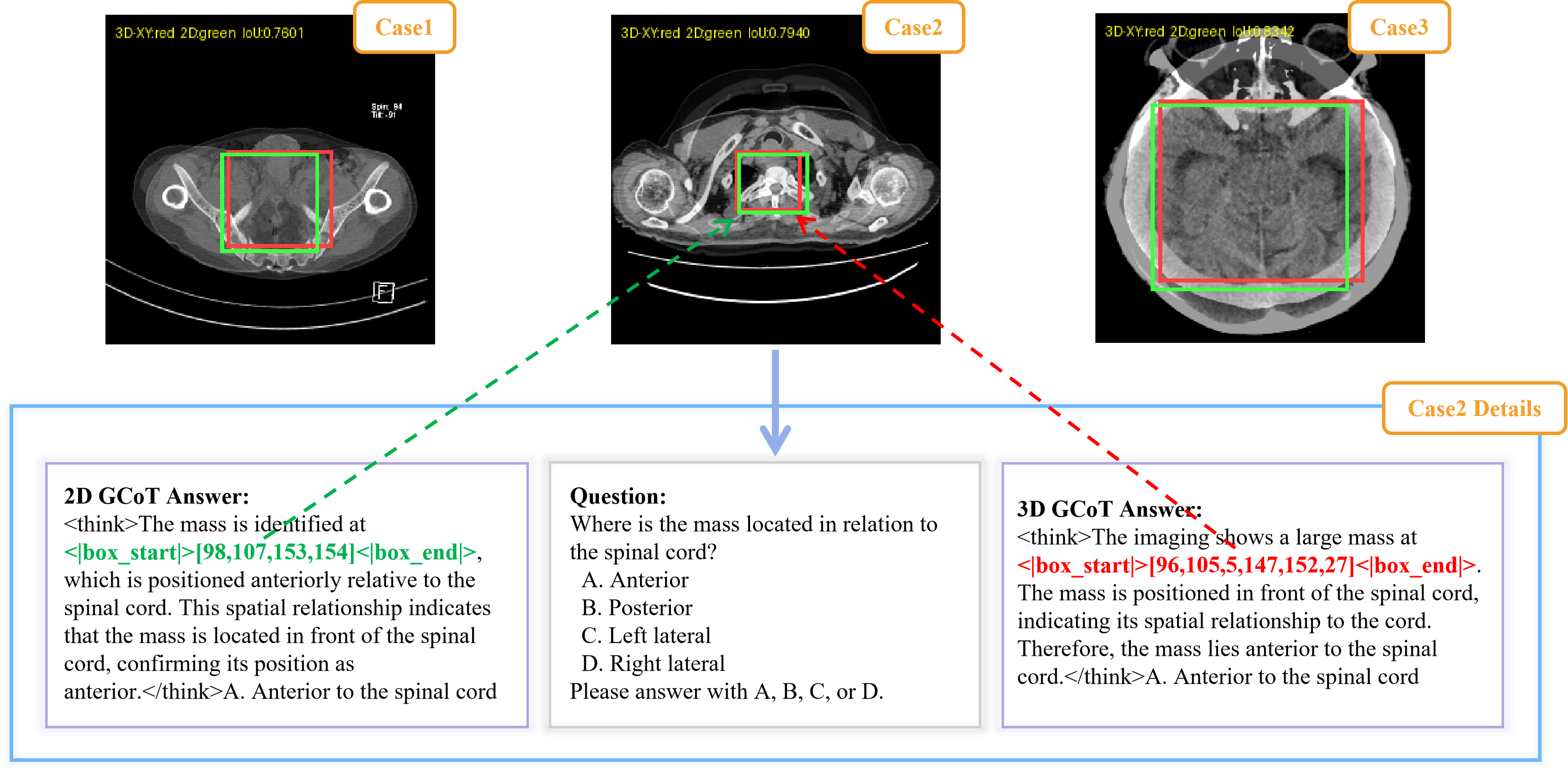} 
    \vspace{-20pt}
    \caption{\textbf{Slice-volume grounding consistency.} The green boxes correspond to slice-wise 2D GCoT predictions, while the red boxes correspond to the $xy$ projection of 3D GCoT cuboids within the predicted slice range. 
    The top part shows three example cases, and the bottom part presents the detailed analysis of Case 2, including the question and the reasoning outputs of both 2D GCoT and 3D GCoT.}
    \label{fig:iou_cases}
    \vspace{-0.3cm}
\end{figure*}

To further understand the optimization process during the Stage-2 Reinforcement Learning, we visualize the training dynamics in Fig.~\ref{fig:training_curve}. The progression of the GRPO stage suggests several training behaviors:

\begin{itemize}
    \setlength{\itemsep}{0pt}
    \item \textbf{Reward Convergence:} As training progresses, the composite reward steadily increases and converges, suggesting gradual improvement of the learned policy under outcome-level supervision.
    \item \textbf{Grounding Precision:} The crop area ratio exhibits a slightly decreasing trend with minor fluctuations. This suggests that the model gradually learns to localize and focus on more precise, relevant visual regions rather than broadly cropping large anatomical areas.
    \item \textbf{Reasoning Depth:} The average response length shows a mild upward trajectory. Instead of finding shortcuts, the model tends to produce more detailed and comprehensive reasoning chains inside the \texttt{<think>} blocks as it optimizes for final answer correctness.
\end{itemize}

These observations suggest that the GRPO framework remains stable while implicitly refining spatial grounding behaviors without requiring step-by-step box-overlap rewards.

\section{Correlation Analysis between Grounding and Reasoning}
\label{sec:appendix_pearson}

To further illustrate the relationship between grounding accuracy and final reasoning performance discussed in Sec.~\ref{sec:experiment}, we plot the correlation between the grounding Intersection-over-Union (IoU) and the answer correctness score across the test set. As illustrated in Fig.~\ref{fig:pearson}, the two metrics exhibit a significant positive correlation, consistent with the hypothesis that more accurate spatial localization tends to provide more reliable visual evidence for medical reasoning.

\section{Qualitative Examples}
\label{sec:appendix_qualitative}

To provide an intuitive view of the slice-volume grounding consistency related to the transfer analysis in RQ1, we compare slice-wise 2D GCoT predictions with the $xy$ projection of 3D GCoT cuboids in Fig.~\ref{fig:iou_cases}. 

The visualization suggests that the shared interface helps the model focus on consistent target anatomical regions across planar slices and slice-serialized volumetric inputs.

\end{document}